\documentclass[conference]{IEEEtran}
\IEEEoverridecommandlockouts
\usepackage{cite}
\usepackage{amsmath,amssymb,amsfonts}
\usepackage{algorithmic}
\usepackage{graphicx}
\usepackage{textcomp}
\usepackage{xcolor}
\usepackage{url}
\usepackage{multirow}
\usepackage{float}
\usepackage{enumitem}
\def\BibTeX{{\rm B\kern-.05em{\sc i\kern-.025em b}\kern-.08em
    T\kern-.1667em\lower.7ex\hbox{E}\kern-.125emX}}
\begin{document}

\title{Machine Unlearning for Class Removal through SISA-based Deep Neural Network Architectures
}

\author{\IEEEauthorblockN{1\textsuperscript{nd} Ishrak Hamim Mahi}
\IEEEauthorblockA{\textit{Department of Computer Science and Engineering} \\
\textit{Brac University}\\
Dhaka, Bangladesh \\
ishrak.hamim.mahi@g.bracu.ac.bd}
\and
\IEEEauthorblockN{2\textsuperscript{st} Siam Ferdous}
\IEEEauthorblockA{\textit{Department of Computer Science and Engineering} \\
\textit{Brac University}\\
Dhaka, Bangladesh \\
siam.ferdous@g.bracu.ac.bd}
\and
\IEEEauthorblockN{3\textsuperscript{rd} Md Sakib Sadman Badhon}
\IEEEauthorblockA{\textit{Department of Computer Science and Engineering} \\
\textit{Brac University}\\
Dhaka, Bangladesh \\
sakib.sadman.badhon@g.bracu.ac.bd}
\and
\IEEEauthorblockN{4\textsuperscript{th} Nabid Hasan Omi}
\IEEEauthorblockA{\textit{Department of Computer Science and Engineering} \\
\textit{Brac University}\\
Dhaka, Bangladesh \\
nabid.hasan.omi@g.bracu.ac.bd}
\and
\IEEEauthorblockN{5\textsuperscript{th} Md Habibun Nabi Hemel}
\IEEEauthorblockA{\textit{Department of Computer Science and Engineering} \\
\textit{Brac University}\\
Dhaka, Bangladesh \\
habibun.nabi.hemel@g.bracu.ac.bd}
\and
\IEEEauthorblockN{6\textsuperscript{th} Dr. Farig Yousuf Sadeque}
\IEEEauthorblockA{\textit{Department of Computer Science and Engineering} \\
\textit{Brac University}\\
Dhaka, Bangladesh \\
farig.sadeque@bracu.ac.bd}
\and
\IEEEauthorblockN{7\textsuperscript{th} Md. Tanzim Reza}
\IEEEauthorblockA{\textit{Department of Computer Science and Engineering} \\
\textit{Brac University}\\
Dhaka, Bangladesh \\
tanzim.reza@bracu.ac.bd}
}

\maketitle

\begin{abstract}
The rapid proliferation of image generation models and other artificial intelligence (AI) systems has intensified concerns regarding data privacy and user consent. As the availability of public datasets declines, major technology companies increasingly rely on proprietary or private user data for model training, raising ethical and legal challenges when users request the deletion of their data after it has influenced a trained model. Machine unlearning seeks to address this issue by enabling the removal of specific data from models without complete retraining. This study investigates a modified SISA (Sharded, Isolated, Sliced, and Aggregated) framework designed to achieve class-level unlearning in Convolutional Neural Network (CNN) architectures. The proposed framework incorporates a reinforced replay mechanism and a gating network to enhance selective forgetting efficiency. Experimental evaluations across multiple image datasets and CNN configurations demonstrate that the modified SISA approach enables effective class unlearning while preserving model performance and reducing retraining overhead. The findings highlight the potential of SISA-based unlearning for deployment in privacy-sensitive AI applications. The implementation is publicly available at \url{https://github.com/SiamFS/sisa-class-unlearning}.

\begin{IEEEkeywords}
Machine Unlearning, Privacy, CNN, AI, SISA, Class Unlearning, Replay Mechanism, Gating Network.
\end{IEEEkeywords}
\end{abstract}

\section{Introduction}

\subsection{Background}
The concept of Artificial Intelligence (AI) can be traced back to early mechanical inventions in ancient Greece, such as Hero of Alexandria’s automata—self-operating mechanisms that laid the conceptual groundwork for automation. Over centuries, advances in logic, mathematics, and computing shaped modern AI. The first practical demonstration appeared in 1951 when Arthur Samuel developed a Checkers Program capable of improving its gameplay through experience \cite{badhon1_Bowling2006Machine}. Around the same time, Alan Turing proposed a chess-playing system that explored the possibility of machine reasoning.

Public attention to AI surged in 1997 when IBM’s Deep Blue defeated world chess champion Garry Kasparov, showcasing AI’s potential and sparking widespread interest \cite{badhon1_Bowling2006Machine}. Subsequent decades witnessed rapid progress in machine learning (ML), especially neural networks, which simulate human cognition through interconnected layers of neurons \cite{lecun2015deep}. Early perceptrons evolved into deep architectures such as Convolutional Neural Networks (CNNs), which extract hierarchical spatial features from visual data \cite{Li2020A}. Although CNNs achieved remarkable success in image recognition, they demanded significant computational resources and large datasets. Enhancements such as batch normalization, residual connections, and the ResNet architecture improved training efficiency and depth \cite{Santos2022Avoiding}.

Further breakthroughs arrived with Generative Adversarial Networks (GANs) and diffusion models \cite{Li2020A}], which enabled high-fidelity image synthesis and manipulation. These generative approaches redefined computer vision and fueled innovations in creative AI applications. Modern systems such as OpenAI’s SORA exemplify the ability of generative AI to produce photorealistic and artistic visual content rivaling human creativity.

\subsection{Motivation}
Large-scale AI models like GPT rely on massive datasets, often scraped from publicly accessible sources. However, this practice raises ethical and legal concerns regarding consent and ownership. With the emergence of data protection laws such as the General Data Protection Regulation (GDPR) and the California Consumer Privacy Act (CCPA), individuals are granted the “right to be forgotten,” compelling organizations to remove user data upon request \cite{omi1_ish1_7163042}.
While data deletion is straightforward in storage systems, removing its influence from trained AI models remains technically challenging. The field of \textbf{machine unlearning} aims to resolve this by developing methods to selectively erase the impact of specific data without retraining entire models. Among various frameworks, the \textbf{SISA (Sharded, Isolated, Sliced, and Aggregated)} framework has shown promise in providing scalable, privacy-preserving unlearning for classification tasks. This research extends the SISA principle to achieve class-level unlearning in CNN architectures, paving the way toward its adaptation in more complex models such as Transformers.

\subsection{Problem Statement}
Unlike databases where records can be explicitly removed, AI models store information in a distributed manner, making targeted data removal highly complex \cite{badhon3_Tax2017The}. The key challenges include:
\begin{itemize}
    \item \textbf{Distributed Representation} - Model knowledge is encoded across numerous parameters, making it difficult to isolate and modify data-specific information.

    \item \textbf{Lack of Explicit Memory} - Neural networks lack identifiable memory slots for individual data points, unlike databases.

    \item \textbf{Hierarchical Feature Learning} - CNN layers intertwine low- and high-level features, complicating selective class removal without performance loss.

    \item \textbf{Lack of Testing Method} - the existing testing methods are ambiguous to ensure that the data has been deleted from the model.

    \item \textbf{Testing Ambiguity} - There are no definitive methods to verify complete data removal.

    \item \textbf{Shared Convolutional Filters} - Filters reused across classes make selective forgetting risk degrading performance on other classes.

    \item \textbf{Black-Box Nature} - It remains opaque how specific data influences overall model behavior, making precise deletion nearly impossible.

    \item \textbf{Class Interdependency} - Removing one class alters decision boundaries, leading to misclassification.

\end{itemize}
Consequently, organizations often resort to retraining from scratch as this approach is computationally expensive and energy-intensive, particularly for large-scale CNN and generative models. This work explores \textbf{efficient class-level unlearning mechanisms} to address these limitations.

\subsection{Research Objectives}
Researchers have developed various methods for machine unlearning, each producing different results depending on whether the goal is to remove individual data points, features, or entire classes. Among these, the \textbf{SISA framework} remains one of the most practical and widely adopted approaches. While the original SISA design focused on removing specific data points from CNN models, this study addresses a broader challenge—\textbf{removing entire classes} from trained CNN architectures using a \textbf{modified SISA framework}.
This form of \textbf{class unlearning} aligns more closely with real-world privacy requirements, where organizations may need to eliminate all data related to a particular category rather than isolated samples. The primary objective of this research is to demonstrate that the modified SISA framework can effectively unlearn full classes from CNN models without requiring complete retraining, thereby offering a \textbf{scalable and practical solution for privacy compliance}.

\section{Related Works}

Machine unlearning is a complex process of making a machine learning model forget certain information more specifically data, which is important for privacy laws like the GDPR \cite{siam3_omi2_hemel1_ish2_9519428}. Regulations like the European Union's GDPR and the California's CCPA require models to remove specific user data if requested in order to enforce the "right to be forgotten" popularly referred to as RTBF \cite{omi1_ish1_7163042}. Conventional model retraining is often impractical due to high computational costs which makes unlearning techniques a legally viable solution \cite{ish7_shokri2017membershipinferenceattacksmachine}. As models grow in size and complexity, full retraining becomes prohibitively expensive \cite{ish6_henighan2020scalinglawsautoregressivegenerative}. A good unlearning algorithm should ensure that the new model behaves almost the same as a retrained one while being much faster \cite{hemel5_10488864}. One of the earliest frameworks for machine unlearning is the SISA (Sharded, Isolated, Sliced and Aggregated) training framework. Introduced by \cite{siam3_omi2_hemel1_ish2_9519428} this framework aims to achieve the unlearning process by strategically limiting the influence of individual data points during model training.

The EMN Framework is a novel machine unlearning framework proposed by \cite{ish10_Tarun_2024}. The EMN stands for Error Maximizing Noise. This framework works by deliberately injecting noise which maximizes error for the class or data which is to be forgotten while retaining the overall performance of the model. SIBU stands for Statistical Inference-Based Unlearning which was introduced by \cite{omi1_ish1_7163042}. This approach uses statistical tools like hypothesis testing or confidence intervals to find how much influence does a data point (which is to be forgotten) have on model decisions. \cite{siam8_kga_wang-etal-2023-kga} propose Knowledge Gap Alignment (KGA), a novel unlearning framework for NLP that is particularly effective for classification, machine translation, and response generation that efficiently removes data from a model while maintaining its performance. \cite{siam4_liu2022continuallearningprivateunlearning} introduces the CLPU-DER++ framework for Continual Learning and Private Unlearning that allows models to sequentially learn and selectively unlearn specific tasks. Gradient-Based Approximate Unlearning (GBAU) is a strategy for rapidly eliminating the influence of certain data points from a trained machine learning model without having to start from scratch \cite{siam8_fan2024salunempoweringmachineunlearning}.

\begin{figure}[h]
    \centering
    \includegraphics[width=\columnwidth]{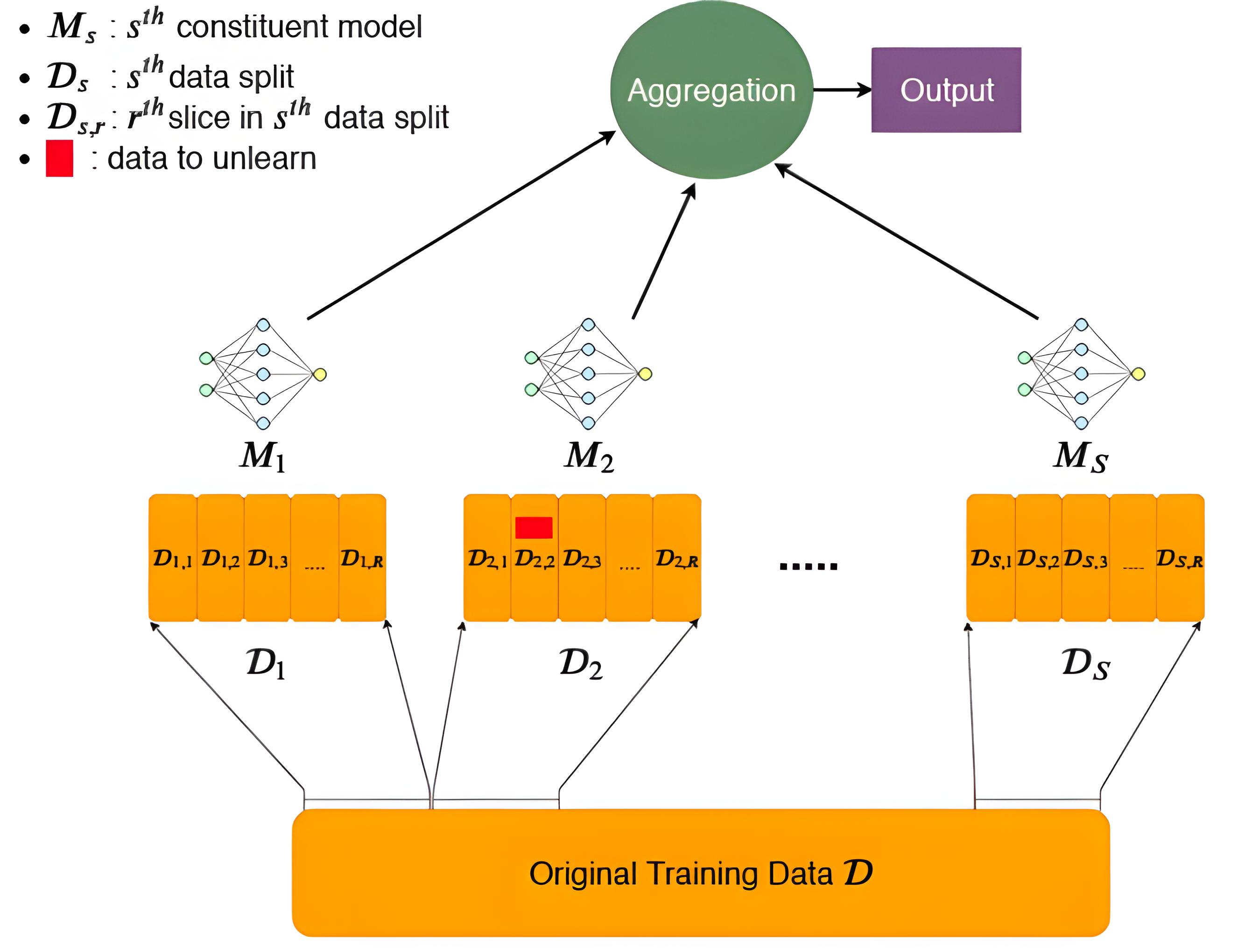}
    \caption{SISA Framework}
    \label{fig:sisa_framework}
\end{figure}

Zero-glance unlearning enables the system to forget information without going over it again. When data deletion laws prevent organizations from keeping a duplicate of their data, this predicament arises \cite{hemel6_Chundawat_2023}. The zero-shot unlearning method is a controlled process where the machine loses some information and it will have no access to the training data \cite{omi5_hemel3_9157084}. Moreover, the old model is adjusted to make it behave like it did not train using the forgotten data \cite{hemel6_Chundawat_2023}. The few-shot unlearning method allows the use of a very small percentage of the erased data to help during the method of unlearning \cite{hemel7_yoon2023fewshotunlearningmodelinversion}.

Exact unlearning assures that the model's parameters are completely restored to a statistically indistinguishable condition as if the removed data never existed. This can be accomplished by utilizing strategies such as re-training from scratch without the to be removed data ($D_r$) sample or efficient checkpointing systems \cite{omi3_ish5_ginart2019makingaiforgetyou}. To overcome the inefficiency of exact unlearning, approximate unlearning strategies seek to eliminate the effects of $D_r$ without requiring total retraining. These methods include influence function-based updates, statistical perturbations, and gradient-based approaches that change model parameters locally \cite{omi5_hemel3_9157084}. The main problem is determining how much residual information from $D_r$ still influences the model and whether it passes privacy requirements \cite{omi6_hemel2_neel2020descenttodeletegradientbasedmethodsmachine}. Unlearning scenarios explain how the machine learning model will forget particular data, it can be a small portion of the full dataset and it can also be a big portion of the main dataset by maintaining the model's performance. The effectiveness, complexities, feasibility, and computational cost of each scenario vary, depending on the system's capacity to handle access to the training data \cite{hemel6_Chundawat_2023}.

\section{Dataset Analysis and Preprocessing}

\begin{figure}[h]
    \centering
    \includegraphics[width=0.8\columnwidth]{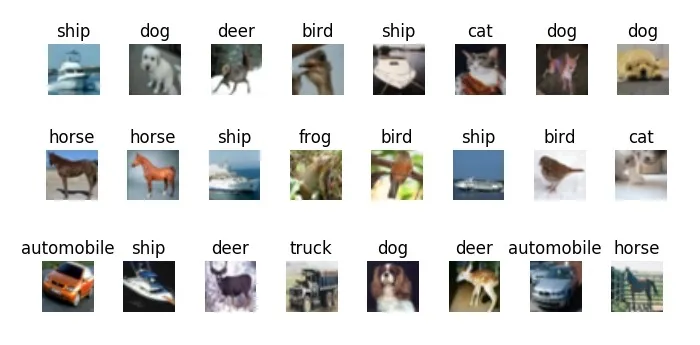}
    \caption{CIFAR-10 Dataset Sample}
    \label{fig:cifar_dataset_sample}
\end{figure}

We selected the CIFAR-10 dataset as the primary benchmark for implementing and evaluating our modified SISA framework for class-level unlearning. CIFAR-10 comprises 60,000 color images (32×32 pixels) distributed across 10 mutually exclusive classes: airplane, automobile, bird, cat, deer, dog, frog, horse, ship, and truck (Figure~\ref{fig:cifar_dataset_sample}). CIFAR-10 provides a critical balance in computer vision benchmarks which are more challenging than MNIST yet computationally accessible unlike ImageNet, enabling rapid experimentation essential for unlearning research.

Our preprocessing pipeline converts images to PyTorch format (HWC to CHW) and applies normalization for stable gradient flow. The SISA framework requires a two-phase data partitioning: sharding through class-based assignment, followed by slicing each shard into sequential portions. We split the 60,000 images using a 70-10-20 ratio (42,000 training, 6,000 validation, 12,000 testing) with stratified sampling.

Sharding divides the training dataset into multiple independent subsets, each trained on its own isolated model. This approach enables selective retraining which means only the affected shard needs updating during unlearning. Leading it to faster retraining, improved scalability, and reduced memory usage. Each class is assigned to exactly one shard using a load-balancing algorithm that minimizes the imbalance ratio (largest shard size / smallest shard size). For CIFAR-10 with two shards, five classes are assigned to each shard (21,000 samples per shard), achieving an imbalance ratio of 1.0. We implemented both two-shard and three-shard configurations to evaluate partition granularity impacts.

Slicing further divides each shard into smaller sequential subsets for incremental training. When a class is removed, only the affected slice and subsequent ones require retraining, significantly reducing computational cost. However, sequential training introduces catastrophic forgetting, where the model loses knowledge from earlier slices. The slicing algorithm divides each shard into equal-sized portions while maintaining class cohesion whenever possible. We experimented with three-slice and five-slice configurations under different sharding conditions, balancing unlearning efficiency with model accuracy and stability.

\begin{figure}[h]
    \centering
\includegraphics[width=0.7\columnwidth]{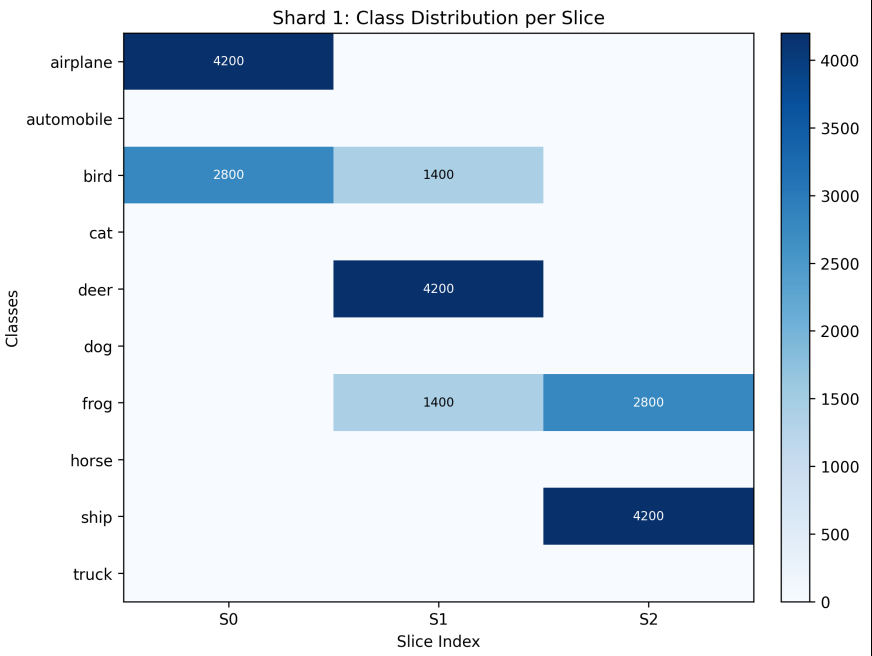}
    \caption{Class Distribution per Slice on Shard 1}
    \label{fig:shard1_slice_distribution}
\end{figure}

Figures~\ref{fig:shard1_slice_distribution} illustrate the sequential filling process for shards. Each shard contains five classes (4,200 samples each) divided into three slices (7,000 samples per slice). The algorithm preserves class grouping while maintaining uniform slice sizes, demonstrating how the approach adapts when slice boundaries do not align perfectly with class boundaries.


\section{Training and Model Development}

This section outlines the progressive development of our models, beginning with a baseline Convolutional Neural Network (CNN) and gradually extending toward increasingly sophisticated variants of the SISA framework. Each stage in this development pipeline was guided by specific limitations identified in the previous stage, leading to methodical architectural and procedural enhancements.

Initially, the baseline CNN served as a reference point for evaluating both classification accuracy and retraining time. However, this approach exhibited clear inefficiencies during unlearning operations: the removal of an entire class required retraining the entire model $\mathcal{M}$ on a reduced dataset $\mathcal{D}_u \setminus \mathcal{C}_x$, resulting in significant computational overhead (see Section~\ref{sec:unlearning_methodology} for a formal unlearning formulation).

To address these inefficiencies, we explored the SISA (Sharded, Isolated, Sliced, and Aggregated) learning paradigm, which partitions the dataset into smaller, structured subsets to localize retraining to specific model components \cite{siam3_omi2_hemel1_ish2_9519428}. Formally, the original dataset $\mathcal{D}_u$ was divided into $K$ disjoint shards $\{\mathcal{S}_1, \mathcal{S}_2, \ldots, \mathcal{S}_K\}$ such that

\begin{equation}
	\mathcal{D}_u = \bigcup_{k=1}^{K} \mathcal{S}_k, \quad \mathcal{S}_i \cap \mathcal{S}_j = \emptyset \quad \forall i \neq j.
	\label{eq:shard_partition}
\end{equation}

Each shard $\mathcal{S}_k$ was further partitioned into slices $\{\mathcal{S}_{k,1}, \mathcal{S}_{k,2}, \ldots, \mathcal{S}_{k,L_k}\}$, and an independent constituent model $\mathcal{M}_k$ was trained sequentially on these slices with checkpointing after each stage.

At each development stage, we jointly present model architecture and training methodology, enabling a structured comparison of how incremental design changes influence learning dynamics, retraining latency, and empirical performance. Corresponding experimental outcomes, including accuracy, training time, and retraining time, are reported in Section~\ref{chap:evaluation}.

\subsection{Baseline Model}
The baseline architecture consists of a single CNN parameterized by $\theta$, denoted $\mathcal{M}_\theta$, trained on the complete dataset $\mathcal{D}_u = \{(x_i, y_i)\}_{i=1}^{m}$ where $y_i \in \mathcal{Y}$ and $|\mathcal{Y}| = C$. This model learns a mapping $\mathcal{M}_\theta : \mathbb{R}^{H\times W \times 3} \to \Delta^{C-1}$ from input images to probability distributions over $C$ classes, where $\Delta^{C-1}$ denotes the probability simplex. The training objective minimizes the categorical cross-entropy loss:
\begin{equation}
	\theta^\ast = \arg\min_{\theta} \mathcal{L}_{\mathrm{CE}}(\mathcal{M}_\theta, \mathcal{D}_u).
\end{equation}

While this baseline provides a reference for classification performance, it presents a significant limitation in the context of machine unlearning. When a class $c^\ast \in \mathcal{Y}$ must be removed, the entire model requires retraining from scratch on the reduced dataset $\mathcal{D}_{u\setminus c^\ast} = \{(x_i,y_i)\in\mathcal{D}_u \mid y_i \ne c^\ast\}$ to obtain new parameters $\theta^\ast_{-c^\ast} = \arg\min_{\theta}\mathcal{L}_{\mathrm{CE}}(\mathcal{M}_\theta, \mathcal{D}_{u\setminus c^\ast})$. This computationally expensive retraining procedure directly motivates the development of partition-based SISA variants, which enable selective updates to affected model components rather than complete retraining. For consistency across all experiments, we employ the Adam optimizer with learning rate $\eta$, apply early stopping based on validation set $\mathcal{D}_{\text{val}}$ performance, and select batch size $B$ to balance training stability and runtime efficiency. Performance evaluation reports both test accuracy and wall-clock training time $T_{\text{train}}$ to facilitate comparative analysis of unlearning efficiency.

\subsection{SISA Framework with Balanced Class Slicing}
To improve retraining efficiency, we partition the dataset $\mathcal{D}_u$ into $K$ shards following the SISA framework, where each shard $\mathcal{S}_k$ is assigned an independent CNN model $\mathcal{M}_{\theta_k}$. The dataset partitioning ensures balanced class distribution across shards, satisfying $P(y \mid x \in \mathcal{S}_k) \approx P(y)$ for all $k$ to prevent shard-specific bias. This produces a set of constituent models $\{ \mathcal{M}_{\theta_1}, \mathcal{M}_{\theta_2}, \ldots, \mathcal{M}_{\theta_K} \}$ trained independently on their respective shards.

During inference, each model generates a probability vector $\mathbf{p}_k(x) = \mathcal{M}_{\theta_k}(x)$ for input $x$, and the final prediction aggregates these outputs via:
\begin{equation}
\hat{y}(x) = \arg\max_{c \in \mathcal{Y}} \max_{k \in \{1, \dots, K\}} \mathbf{p}_k^{(c)}(x),
\end{equation}
where the shard with highest confidence determines the class label. This sharded design provides crucial efficiency advantages for unlearning. When removing class $c^\ast$, only the affected shard $\mathcal{S}_k$ containing $c^\ast$ requires retraining on the reduced dataset $\mathcal{S}_{k \setminus c^\ast}$:
\begin{equation}
\theta_k^\ast \leftarrow \arg\min_{\theta_k} \mathcal{L}_{\mathrm{CE}}(\mathcal{M}_{\theta_k}, \mathcal{S}_{k \setminus c^\ast}),
\end{equation}
while all other shard models remain unchanged. This localized retraining significantly reduces computational cost compared to baseline full retraining. Although partitioning slightly reduces generalization capacity as each model accesses only a data fraction, it achieves substantial gains in retraining efficiency, representing a key advancement in unlearning-oriented architecture design. All models employ consistent training configuration: Adam optimizer with learning rate $\eta$, categorical cross-entropy loss, and identical batch sizes across shards.

\begin{figure}[h]
    \centering
\includegraphics[width=0.7\linewidth]{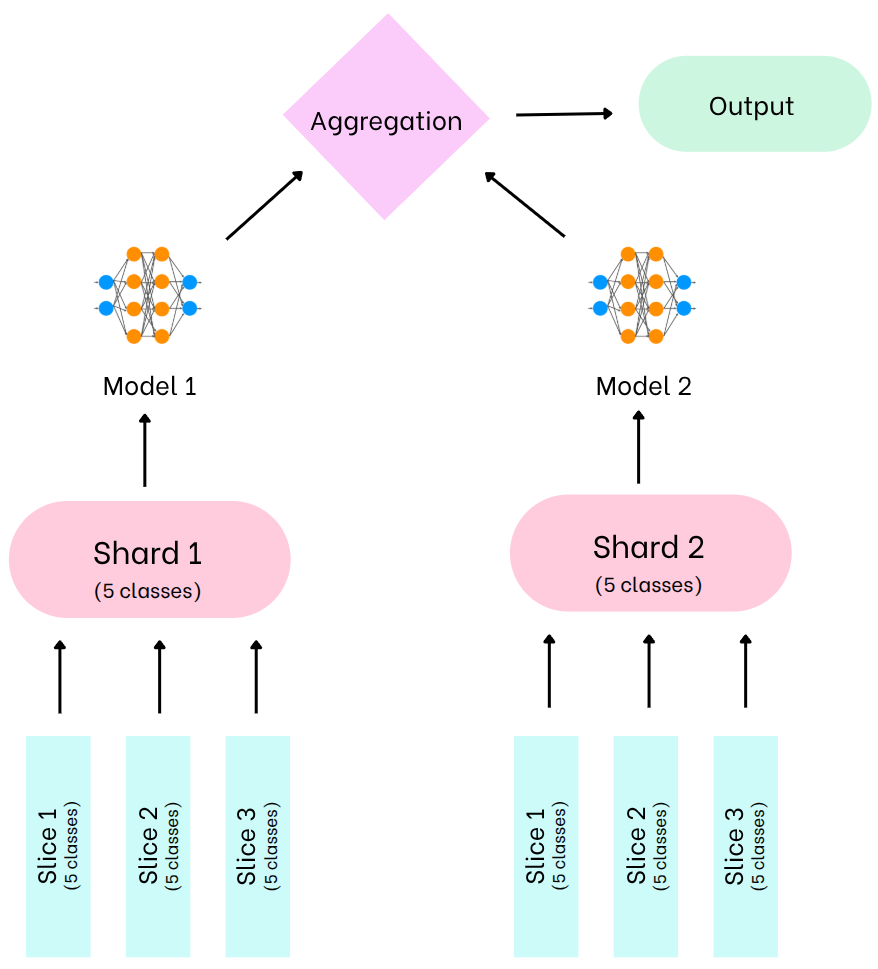}
    \caption{Balanced Class Slicing Model Architecture}
    \label{Balanced Class Slicing Model Architecture}
\end{figure}

\subsection{SISA Framework with Sequential Class Slicing and Replay Mechanism}

This framework extends shard-based partitioning by introducing slice-level isolation. Each shard $\mathcal{D}^{(s)}$ is further partitioned into $L^{(s)}$ sequential slices $\mathcal{D}^{(s)}_\ell$ containing disjoint class subsets:
\begin{equation}
\mathcal{D}^{(s)} = \bigcup_{\ell=1}^{L^{(s)}} \mathcal{D}^{(s)}_\ell, \quad \mathcal{D}^{(s)}_\ell \cap \mathcal{D}^{(s)}_{\ell'} = \emptyset \quad \text{for} \quad \ell \neq \ell'.
\end{equation}

Each shard model $M^{(s)}$ trains incrementally across slices: $M^{(s)}_\ell = \text{Train}( M^{(s)}_{\ell-1}, \mathcal{D}^{(s)}_\ell )$, with parameters checkpointed after each slice ($\theta^{(s)}_\ell = \text{Checkpoint}(M^{(s)}_\ell)$) to enable efficient rollback during unlearning. Inference aggregates shard predictions via $\hat{y} = \arg\max_{y} \max_{s} \, p^{(s)}(y \mid x)$.

However, sequential training on disjoint class slices causes catastrophic forgetting where accuracy on earlier slices $\mathcal{A}_{\leq n-1}$ degrades as training progresses to later slices.

\begin{figure}[h]
    \centering
\includegraphics[width=0.7\linewidth]{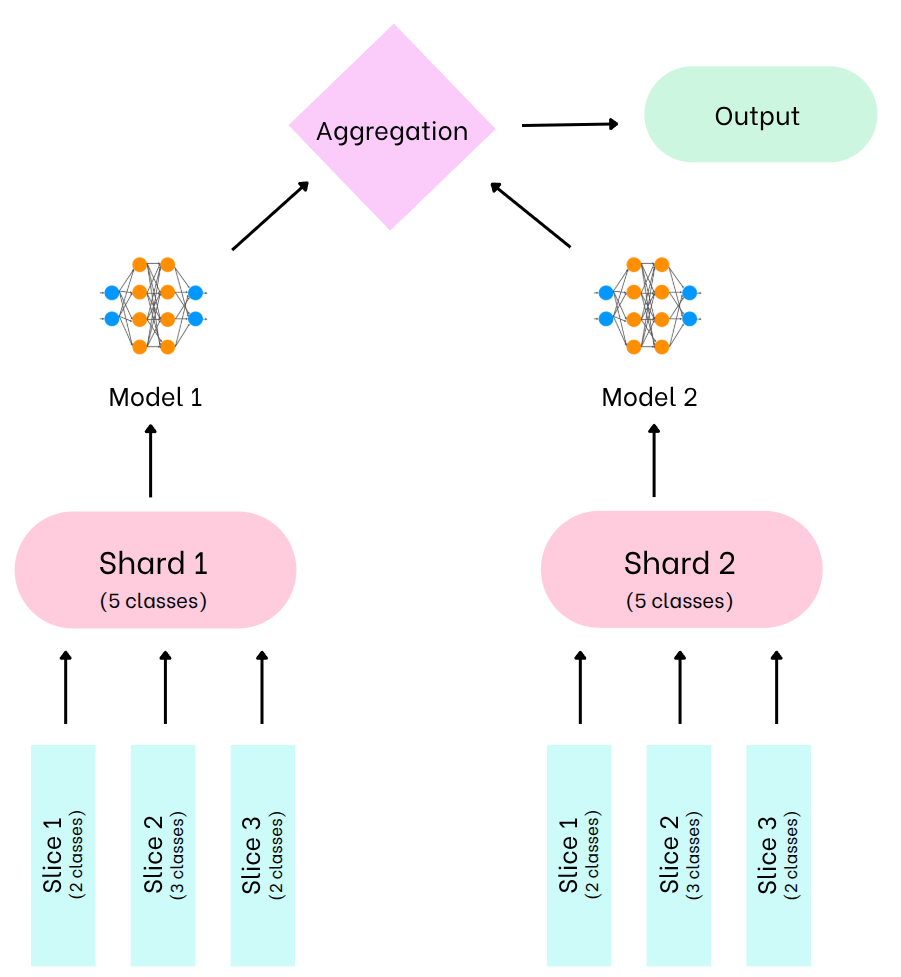}
    \caption{Sequential Class Slicing Model Architecture}
    \label{Sequential Class Slicing Model Architecture}
\end{figure}

To mitigate catastrophic forgetting, we incorporate a replay mechanism where each slice $\ell > 1$ trains on both current data and a subset of previous slices. For slice $\ell$ in shard $s$, a replay subset $\mathcal{R}^{(s)}_\ell \subseteq \bigcup_{j=1}^{\ell-1} \mathcal{D}^{(s)}_j$ is sampled with size determined by replay ratio $\rho$: $|\mathcal{R}^{(s)}_\ell| = \rho \cdot \sum_{j=1}^{\ell-1} |\mathcal{D}^{(s)}_j|$. Training minimizes:
\begin{equation}
\theta^{(s)}_\ell = \arg\min_{\theta} \, \mathcal{L}\Big(\theta; \mathcal{D}^{(s)}_\ell \cup \mathcal{R}^{(s)}_\ell \Big),
\end{equation}
where $\mathcal{L}$ is the cross-entropy loss. Checkpointing after each slice enables efficient unlearning rollback.

\begin{table}[h]
\caption{Performance Comparison of Different Replay Ratios (2 Shards, 5 Slices)}
\label{tab:replay_comparison}
\centering
\begin{tabular}{|c|c|c|c|}
\hline
\textbf{Replay Ratio} & \textbf{Accuracy} & \textbf{Unlearning Acc.} & \textbf{Training Time (s)} \\
\hline
20\% & 69.7\% & 67.8\% & 52.7 \\
\hline
\textbf{30\%} & \textbf{73.1\%} & \textbf{70.7\%} & \textbf{55.4} \\
\hline
40\% & 73.9\% & 71.1\% & 58.2 \\
\hline
\end{tabular}
\end{table}

We evaluated multiple replay ratios on our two-shard, five-slice configuration. As shown in Table~\ref{tab:replay_comparison}, 30\% replay achieves 73.1\% accuracy with 55.4s training time, an optimal balance between forgetting mitigation and computational cost. While 40\% replay yields marginally higher accuracy (73.9\%), the additional training time does not justify the modest 0.8\% improvement. The 20\% ratio, though faster, results in substantially lower accuracy (69.7\%). Therefore, we adopt 30\% replay for all subsequent experiments. This replay-augmented training effectively mitigates catastrophic forgetting while maintaining the retraining efficiency of sequential slicing.

\begin{figure}[h]
    \centering
\includegraphics[width=0.7\linewidth]{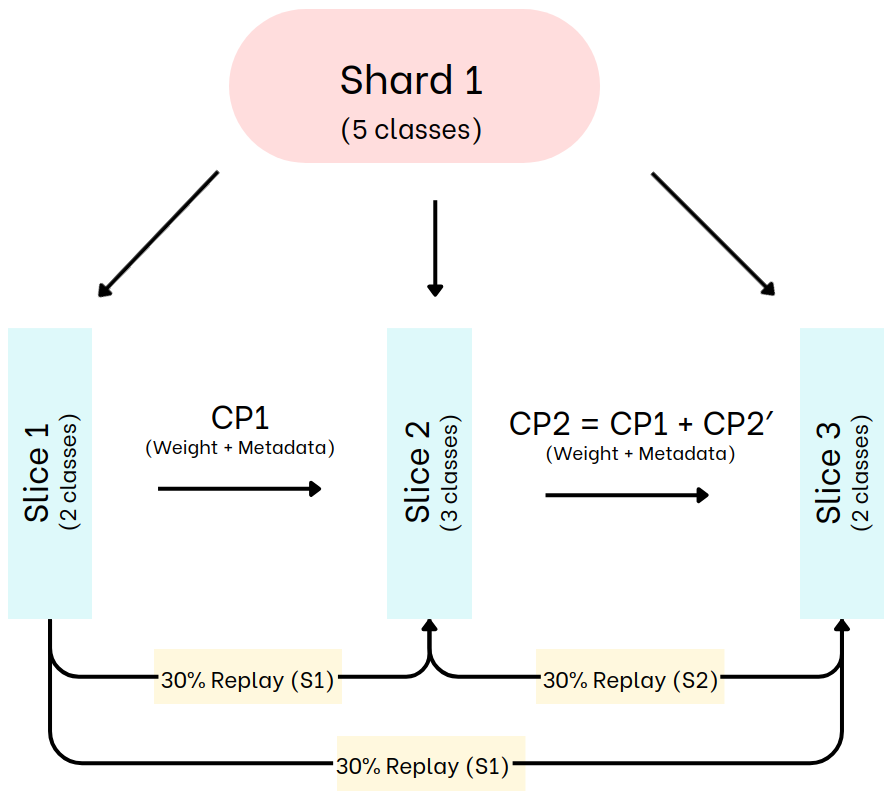}
    \caption{Replay and Checkpointing Mechanism}
    \label{Replay and Checkpointing Mechanism}
\end{figure}

\subsection{SISA Framework with Gating Network}

The final framework variant incorporates a gating network $G$ to route inputs to specialized shard models $\{ M^{(s)} \}_{s=1}^S$, improving prediction accuracy while reducing inference cost. Each shard model is trained using replay-augmented sequential slicing as described previously. The gating network maps inputs $x$ to shard indices by maximizing $p_\phi(s \mid x)$, trained via cross-entropy loss:
\begin{equation}
\mathcal{L}_G(\phi) = - \frac{1}{|\mathcal{D}|} \sum_{(x, s) \in \mathcal{D}} \log p_\phi(s \mid x),
\end{equation}
where shard identifiers $s$ serve as training targets.

During inference, the gating network first selects shard $\hat{s} = \arg\max_{s} \, p_\phi(s \mid x)$, then the selected model $M^{(\hat{s})}$ produces the class prediction $\hat{y} = \arg\max_y \mathbf{p}^{(\hat{s})}(y \mid x)$. This two-stage process directs each input to the most appropriate specialized shard, improving accuracy while activating only one shard model instead of all models. Training proceeds in two stages: first, shard-specific CNNs are trained with replay-augmented slicing and checkpointing; second, the gating network is trained on shard identifiers using consistent preprocessing and optimization. Although the gating network adds modest training overhead, it significantly enhances both accuracy and inference efficiency.

\begin{figure}[h]
    \centering
\includegraphics[width=0.7\linewidth]{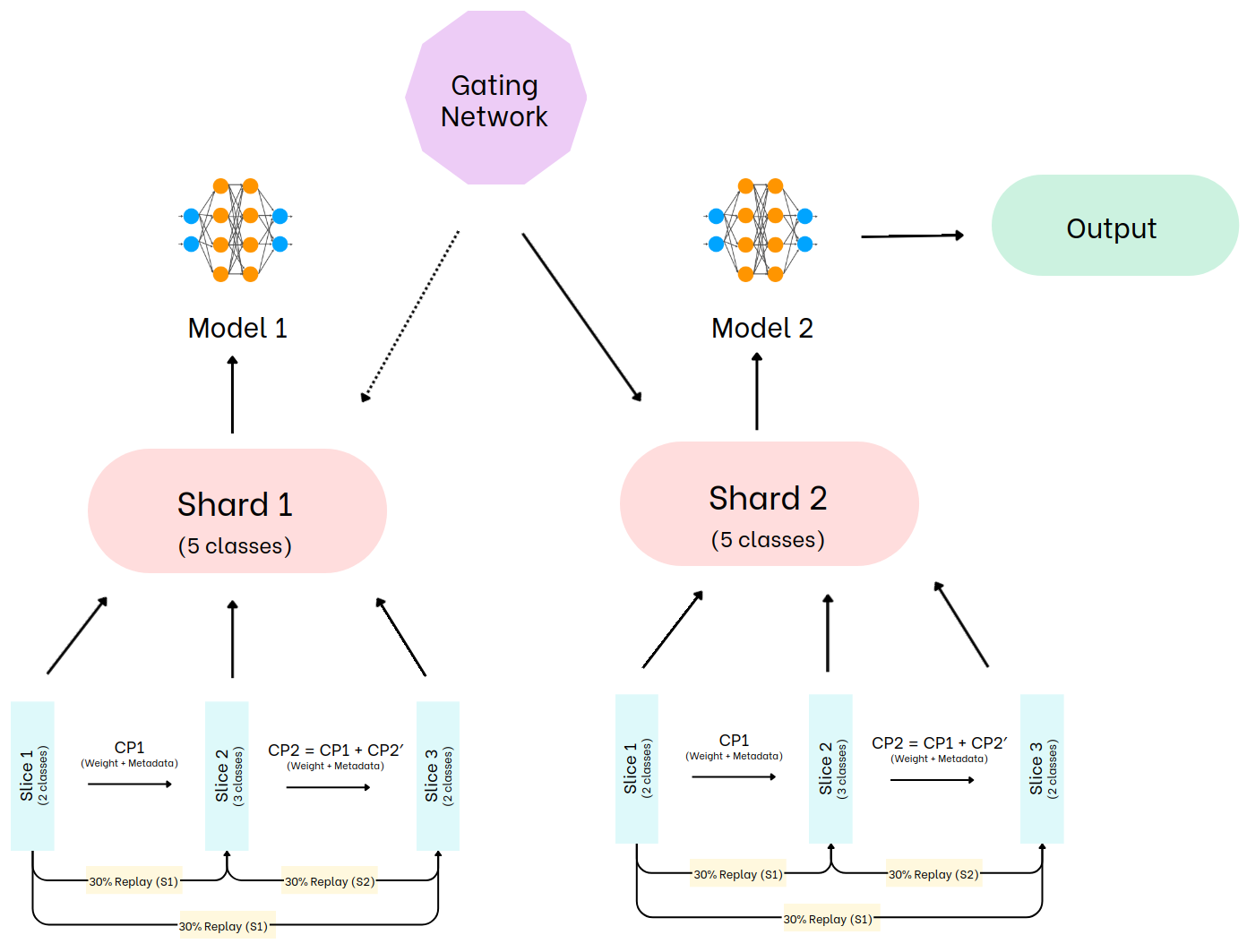}
    \caption{Full Model Architecture with Gating Network}
    \label{Full Model Architecture with Gating Network}
\end{figure}

\section{Unlearning Methodology and Evolution}
\label{sec:unlearning_methodology}

This section presents our class-level unlearning methodology using a modified SISA architecture. While the original SISA framework focused on individual data point removal, we extend this to enable entire class removal, a scenario more aligned with real-world applications. We progressively develop from a baseline model through increasingly sophisticated SISA variants, integrating replay mechanisms and gating networks to enhance post-unlearning performance. The primary objective is minimizing retraining time while maintaining model accuracy. Our approach partitions the dataset into shards, each training an independent constituent model. Upon an unlearning request, the system identifies the shard containing the target class, removes it from the corresponding slices, and retrains only the affected constituent model. Since the target class is completely eliminated, this ensures exact unlearning, validated through confusion matrix analysis presented in the evaluation section.

\subsection{Baseline Model: Full Retraining Approach}
Our baseline architecture consists of a standard Convolutional Neural Network (CNN) trained on the complete dataset without any sharding or structural modifications. Let the initial dataset be denoted as per Equation~\eqref{eq:shard_partition}:

Let $\mathcal{M}_\theta$ denote the CNN model parameterized by $\theta$. Initially, the model is trained on as per Equation~\eqref{eq:shard_partition}, which can be represented as
\begin{equation}
\theta^\ast = \arg\min_{\theta} \mathcal{L}(\mathcal{M}_\theta, \mathcal{D}_u),
\end{equation}
where $\mathcal{L}$ is the empirical loss function, typically cross-entropy for classification tasks. We denote the trained model as $\mathcal{M}(\mathcal{D}_u)$.

Suppose an unlearning request is received to forget a specific class $c \in \mathcal{Y}$. The baseline approach first scans through the entire dataset to identify and remove all samples belonging to class $c$:
\begin{equation}
\mathcal{D}_{u \setminus c} = \{(x_i, y_i) \in \mathcal{D}_u \mid y_i \neq c \}.
\end{equation}
After this filtering, the model is fully retrained from scratch on the reduced dataset $\mathcal{D}_{u \setminus c}$:
\begin{equation}
\theta^\ast_{-c} = \arg\min_{\theta} \mathcal{L}(\mathcal{M}_\theta, \mathcal{D}_{u \setminus c}),
\end{equation}
resulting in a new model $\mathcal{M}(\mathcal{D}_{u \setminus c})$ that has no exposure to class $c$.

In terms of training time, this baseline method requires
\begin{equation}
T_{\text{unlearn}}^{\text{baseline}} \approx T_{\text{train}}(\mathcal{D}_u),
\end{equation}
since the model is retrained on almost the entire dataset, and the computational cost is proportional to the dataset size. Similarly, the accuracy of the baseline model before and after unlearning can be expressed as
\begin{equation}
\text{Acc}(\mathcal{M}(\mathcal{D}_u)) \geq \text{Acc}(\mathcal{M}(\mathcal{D}_{u \setminus c})),
\end{equation}
where the inequality reflects the expected performance degradation due to both the reduced training set size and the removal of class $c$.

This baseline serves as a reference point for evaluating the efficiency and accuracy trade-offs of our proposed unlearning methods.

\subsection{SISA Framework with Balanced Class Slicing}
To reduce the retraining time associated with full model retraining, we implement the SISA (Sharded, Isolated, Sliced, and Aggregated) framework, which partitions the dataset into multiple shards and slices, and introduces intermediate checkpoints at the slice level to further minimize the computational cost during unlearning.

As per Equation~\eqref{eq:shard_partition}, each shard is further divided into $L$ slices:

\begin{equation}
\mathcal{S}_k = \bigcup_{\ell=1}^{L} \mathcal{S}_{k,\ell}, \quad \mathcal{S}_{k,\ell_1} \cap \mathcal{S}_{k,\ell_2} = \emptyset \ \ \forall \ell_1 \neq \ell_2.
\end{equation}
For each shard $\mathcal{S}_k$, a constituent model $\mathcal{M}_k$ is trained sequentially slice by slice:
\begin{equation}
\theta_{k,\ell}^\ast = \arg\min_{\theta} \mathcal{L}(\mathcal{M}_\theta, \mathcal{S}_{k,\ell}), \quad \ell = 1, \ldots, L,
\end{equation}
with checkpoints saved after each slice to allow partial retraining in the event of unlearning.

Let $c \in \mathcal{Y}$ denote the class to be unlearned. Upon receiving an unlearning request, the system consults a metadata table $\mathcal{M}_\text{meta}$ that stores the class-to-shard mapping:
\begin{equation}
\mathcal{M}_\text{meta}(c) \mapsto k^\ast,
\end{equation}
where $k^\ast$ is the shard index containing samples of class $c$. Within shard $\mathcal{S}_{k^\ast}$, a linear scan is performed to locate and remove all samples associated with class $c$:
\begin{equation}
\mathcal{S}_{k^\ast,\ell}^{-c} = \{(x_i, y_i) \in \mathcal{S}_{k^\ast,\ell} \mid y_i \neq c \}, \quad \forall \ell \in \{1,\ldots,L\}.
\end{equation}

In the original SISA framework, data points are distributed among slices in a balanced fashion to facilitate fine-grained unlearning of individual samples. Formally, if $n_c$ denotes the number of samples of class $c$, then under balanced slicing each slice contains approximately $\frac{n_c}{L}$ samples of class $c$. While this is efficient for single-sample unlearning, it introduces significant overhead for class-level unlearning, since the removal of class $c$ requires modifications to all slices within shard $k^\ast$:
\begin{equation}
\text{Slices to retrain for class } c = L.
\end{equation}
This effectively negates some of the time-saving advantages of slicing, as each affected slice must be purged and retrained.

After the constituent model $\mathcal{M}_{k^\ast}$ is retrained on the modified slices
\begin{equation}
\mathcal{S}_{k^\ast}^{-c} = \bigcup_{\ell=1}^L \mathcal{S}_{k^\ast,\ell}^{-c},
\end{equation}
the overall model resumes inference. For evaluation, each input sample is forwarded to all $K$ constituent models $\{\mathcal{M}_1, \ldots, \mathcal{M}_K\}$. The outputs are then aggregated through an aggregation layer, which selects the final prediction via the $\mathrm{argmax}$ function over the class probability distributions:
\begin{equation}
\hat{y} = \arg\max_{y \in \mathcal{Y}} \left( \sum_{k=1}^{K} p_k(y \mid x) \right),
\end{equation}
where $p_k(y \mid x)$ is the predicted probability of class $y$ from constituent model $\mathcal{M}_k$.

\subsection{SISA Framework with Sequential Class Slicing and Replay Mechanism}

While balanced slicing improved retraining efficiency compared to the baseline, it was not optimized for class-level unlearning. The original SISA design distributes samples of each class uniformly across all slices, which benefits single data point removal but requires modifying every slice when removing an entire class. To address this inefficiency, we introduce Sequential Class-Level Slicing (SCLS), which assigns distinct classes to different slices within each shard:
\begin{equation}
\mathcal{S}_k = \bigcup_{\ell=1}^{L} \mathcal{S}_{k,\ell},
\quad \text{with} \quad
\mathcal{Y}_{k,\ell} \cap \mathcal{Y}_{k,\ell'} = \emptyset \quad \forall \, \ell \neq \ell',
\end{equation}
where $\mathcal{Y}_{k,\ell}$ denotes class labels in slice $\ell$ of shard $k$. Under SCLS, when unlearning class $c$, the metadata table identifies both shard $k^\ast$ and slice $\ell^\ast$ containing $c$: $\mathcal{M}_{\text{meta}}(c) \mapsto (k^\ast, \ell^\ast)$. The targeted slice is deleted, and model $\mathcal{M}_{k^\ast}$ is partially retrained from the previous checkpoint on subsequent slices only, requiring $L - \ell^\ast + 1$ slice retraining versus $L$ slices in the balanced case.

However, sequential training on disjoint class slices causes catastrophic forgetting, the model overwrites earlier representations as training progresses. To mitigate this, we integrate a replay mechanism where each slice $n$ trains on both current data $\mathcal{S}_{k,n}$ and a replay buffer $\mathcal{R}_{k,n}$ sampled from previous slices: $\widetilde{\mathcal{S}}_{k,n} = \mathcal{S}_{k,n} \cup \mathcal{R}_{k,n}$. The replay buffer contains $\rho = 0.3$ (30\%) of important samples from all previous slices: $|\mathcal{R}_{k,n}| \approx \rho \times \sum_{j=1}^{n-1} |\mathcal{S}_{k,j}|$. The training objective becomes $\theta_{k,n}^\ast = \arg\min_{\theta} \mathcal{L}(\mathcal{M}_\theta, \widetilde{\mathcal{S}}_{k,n})$, jointly optimizing over current and replayed distributions.

\subsection{SISA Framework with Gating Network}

To further improve prediction accuracy and optimize inference efficiency, we incorporate a lightweight gating network $\mathcal{G}_\phi$ that routes inputs to the most relevant constituent model. The gating network transforms class labels $y_i \in \mathcal{Y}$ into shard identifiers $s_i \in \{1,\ldots,K\}$ via mapping $g: \mathcal{Y} \rightarrow \{1, \ldots, K\}$, and is trained to predict shard labels:
\begin{equation}
\phi^\ast = \arg\min_{\phi} \, \mathcal{L}_{\text{gate}}\!\left( \mathcal{G}_\phi(x_i), s_i \right),
\end{equation}
where $\mathcal{L}_{\text{gate}}$ is cross-entropy loss over shard labels. Since the gating network trains only on shard IDs rather than class labels, it maintains data isolation and unlearning compliance.

During inference, the gating network selects the most probable shard $s^\ast = \arg\max_{s} p_\phi(s \mid x)$, where $p_\phi(s \mid x) = \text{softmax}( \mathcal{G}_\phi(x) )$. The input is then forwarded to the selected constituent model $\mathcal{M}_{s^\ast}$, producing the final prediction $\hat{y} = \arg\max_{y \in \mathcal{Y}} p_{s^\ast}(y \mid x)$. This routing mechanism improves accuracy by directing inputs to specialized shards while reducing inference overhead, querying only one constituent model instead of all $K$ models. Empirically, we observe approximately 10\% accuracy improvement compared to the replay-based version. The unlearning mechanism remains unchanged: class-level unlearning operates at the shard and slice level, while the gating network remains unaffected since it stores only shard-level mappings. The gating network contains 10-15\% of the total parameters across all constituent models, making it a lightweight, tunable component that balances routing accuracy and computational cost.

\section{Result Analysis and Evaluation}
\label{chap:evaluation}

This section presents a detailed evaluation of all four developed models across four different shard–slice configurations to assess their training performance, unlearning efficiency, and retraining behavior. The configurations tested were (2 shards slices), (2 shards 5slices), (3shards 3slices), and (3shards 5slices). For each configuration, we measured four key metrics: Accuracy before unlearning, Training time before unlearning, Accuracy after unlearning, and Average retraining time after unlearning.
An early stopping mechanism was applied during training with a patience of 7 steps, meaning training stopped automatically if the validation loss failed to improve for seven consecutive checks. 

Across all configurations, there was a clear and consistent trade-off between model accuracy and training/retraining efficiency. As the number of shards and slices increased, models trained faster but achieved lower overall accuracy. This is expected because dividing the dataset into smaller shards limits the amount of data each model sees, slightly restricting generalization. However, this same structure provides the crucial advantage of selective retraining, where only specific shards or slices need to be updated when unlearning a class.

After unlearning, all models exhibited some degree of accuracy drop. This drop was not a failure of learning but a direct confirmation of exact unlearning, since the test dataset still contained samples from the deleted class, the model could no longer predict them, automatically lowering overall accuracy. This validates that the model truly forgot the targeted class information.

The table below shows the accuracy and training time of before unlearning for all of our architecture versions for all the shard-slice setups. This table also shows the accuracy and average retraining time after unlearning for all the architecture versions for all the shard-slice setups. In this table the Baseline Model is being represented as Architecture 1 and similarly SISA Balanced Class Distribution as Model 2, SISA with Sequential Class Distribution and Replay Mechanism as Model 3 and SISA with Replay and Gating as Model 4.

\subsection{Accuracy and Time Performance Overview}

Across all configurations, there was a clear trade-off between model accuracy and training/retraining efficiency. As the number of shards and slices increased, models trained faster but achieved lower overall accuracy, as dividing the dataset limits generalization. However, this structure provides the crucial advantage of selective retraining during unlearning operations.

After unlearning, all models exhibited accuracy drops, which directly confirms exact unlearning, since the test dataset still contained samples from the deleted class, the model could no longer predict them, automatically lowering overall accuracy.

\subsection{Model-wise Evaluation}

\subsubsection{Model 1: Baseline CNN}
The baseline model achieved 81.67\% validation accuracy before unlearning and 75.78\% after unlearning. It served as the reference point with highest accuracy but required nearly full retraining time for unlearning. Figure shows the confusion matrix confirming exact unlearning, with the "dog" class showing zero predictions.

\begin{figure}[h]
    \centering
\includegraphics[width=0.6\columnwidth]{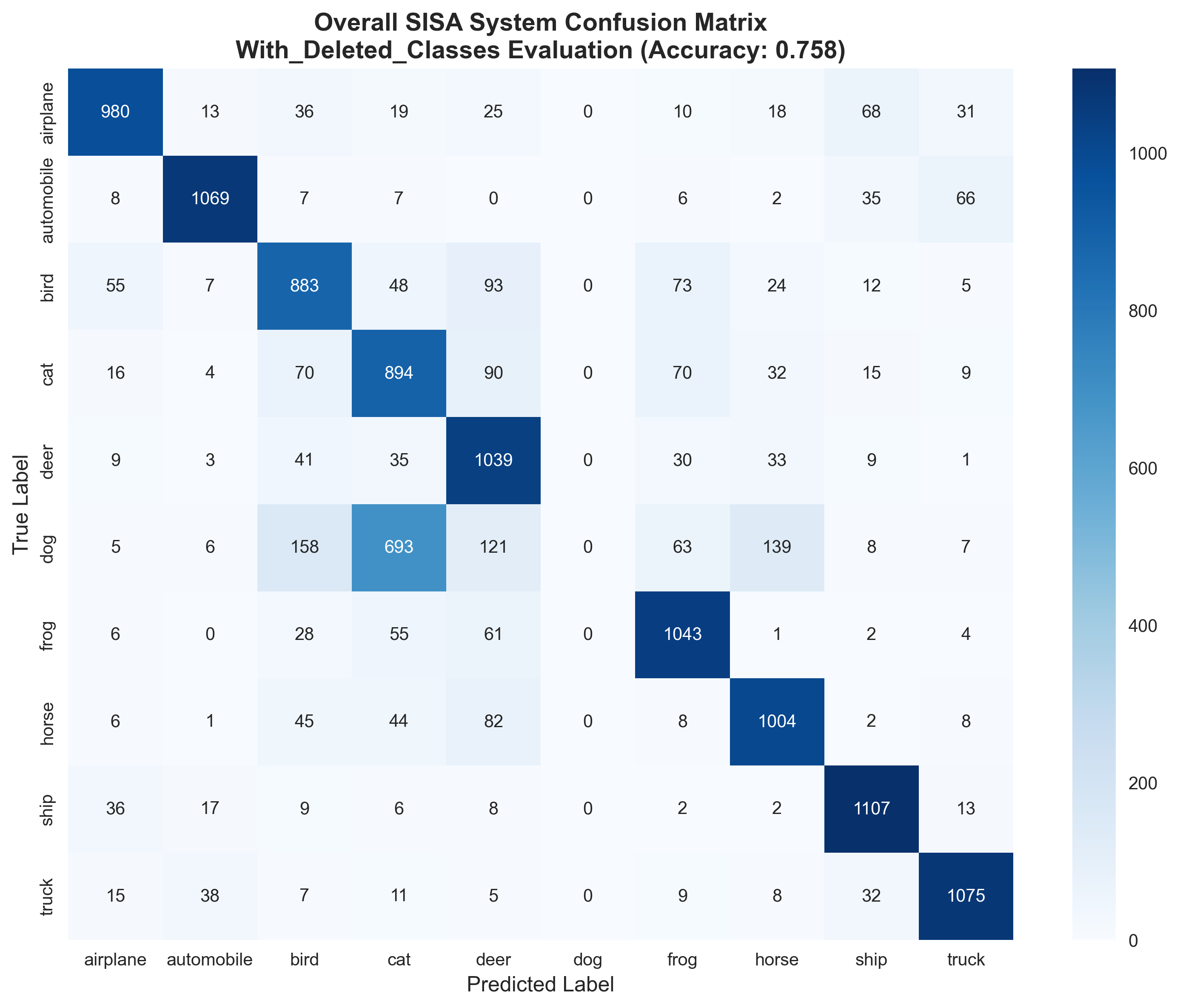}
    \caption{Confusion Matrix after Unlearning "Dog" Class for Baseline CNN}
    \label{fig:confusion_matrix_baseline}
\end{figure}

\subsubsection{Model 2: SISA without Slice Isolation}
By dividing the dataset into shards, this model achieved faster retraining. Each shard-specific CNN was trained independently, and during unlearning, only the shard containing the deleted class needed retraining. This version achieved 67.21\% before unlearning accuracy in the 2×3 setup and 59.99\% after unlearning.

\subsubsection{Model 3: SISA with Sequential Class Slicing and Replay Mechanism}
After adding the sequential class slicing the model suffered for catastrophic forgetting. To address forgetting, a 30\% replay mechanism was added. This version achieved 66.19\% before unlearning accuracy and 61.39\% after unlearning. The replay mechanism slightly increased training time but substantially improved performance.

\subsubsection{Model 4: SISA with Gating Network}
\begin{figure}[h]
    \centering
\includegraphics[width=0.6\columnwidth]{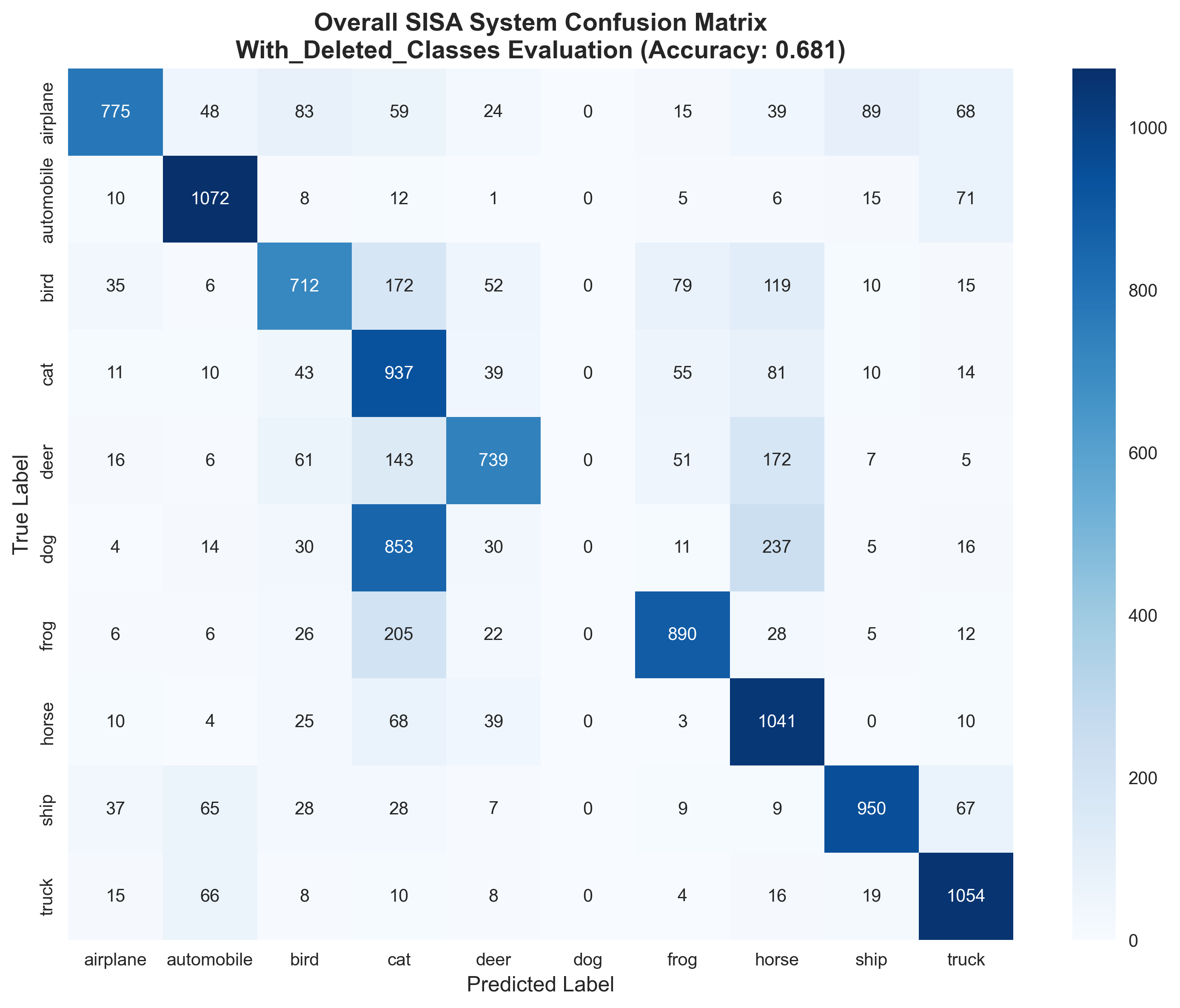}
    \caption{Confusion Matrix with Deleted "Dog" Class in Final Model}
    \label{fig:confusion_final}
\end{figure}

The final model incorporated a gating network for routing inputs to appropriate shards. This achieved 73.12\% before unlearning and 70.12\% after unlearning—the closest to baseline performance, trailing by only 8.55\% before unlearning and 5.66\% after unlearning. The confusion matrix (Figure~\ref{fig:confusion_final}) confirms exact unlearning with zero predictions for the deleted "dog" class.

\subsection{Configuration-Based Analysis}

When comparing across shard-slice setups, more shards reduced training and retraining time but slightly lowered accuracy. More slices further decreased retraining time by limiting update scope, but increased forgetting risk. The replay mechanism largely countered this, while gating ensured accurate routing despite shard specialization.

\subsection{Quantitative Results}

Table~\ref{tab:performance} summarizes performance across all configurations. Model 4 consistently achieved the best balance between accuracy and retraining efficiency.

\begin{table}[h]
\centering
\caption{Performance Across Shard-Slice Configurations}
\label{tab:performance}
\begin{tabular}{|c|c|c|c|c|c|}
\hline
\multicolumn{2}{|c|}{} & \multicolumn{2}{c|}{Before Unlearning} & \multicolumn{2}{c|}{After Unlearning} \\
\hline
Setup & Mdl & Acc\% & T.Time(s) & A.Acc\% & A.RT(s) \\
\hline
\multirow{4}{*}{2-3} & 1 & 81.67 & 77.79 & 75.78 & 69.86 \\
\cline{2-6}
 & 2 & 67.21 & 44.34 & 59.99 & 18.71 \\
\cline{2-6}
 & 3 & 66.19 & 57.62 & 61.39 & 16.66 \\
\cline{2-6}
 & 4 & \textbf{73.12} & 95.53 & \textbf{70.12} & 17.33 \\
\hline
\multirow{4}{*}{2-5} & 1 & 81.67 & 77.79 & 75.78 & 69.86 \\
\cline{2-6}
 & 2 & 64.31 & 39.03 & 57.22 & 18.1 \\
\cline{2-6}
 & 3 & 64.17 & 63.74 & 59.47 & 15.53 \\
\cline{2-6}
 & 4 & \textbf{72.55} & 93.11 & \textbf{70.17} & 16.1 \\
\hline
\multirow{4}{*}{3-3} & 1 & 81.67 & 77.79 & 75.78 & 69.86 \\
\cline{2-6}
 & 2 & 56.23 & 37.53 & 48.73 & 10.23 \\
\cline{2-6}
 & 3 & 53.33 & 47.3 & 46.47 & 15.1 \\
\cline{2-6}
 & 4 & \textbf{72.42} & 96.18 & \textbf{69.98} & 11.2 \\
\hline
\multirow{4}{*}{3-5} & 1 & 81.67 & 77.79 & 75.78 & 69.86 \\
\cline{2-6}
 & 2 & 54.88 & 41.47 & 48.51 & 10.01 \\
\cline{2-6}
 & 3 & 52.4 & 53.4 & 45.8 & 14.3 \\
\cline{2-6}
 & 4 & \textbf{71.83} & 91.8 & \textbf{68.01} & 10.59 \\
\hline
\end{tabular}
\end{table}

\subsection{Summary of Findings}

The evolution from simple sharding to slice isolation, replay, and gating led to steady improvements in both learning stability and retraining efficiency. Sequential class distribution introduced computational efficiency but lacked accuracy due to catastrophic forgetting. The 30\% replay mechanism incorporated variance while maintaining low retraining time, significantly boosting performance. While the baseline CNN remains unmatched in raw accuracy, Model 5 achieves near-equivalent performance with a fraction of the retraining cost and demonstrates true exact unlearning. This represents an effective balance between computational efficiency, data privacy compliance, and accuracy for real-world unlearning systems.

\section{Conclusion}

This research investigated the implementation of machine unlearning using the SISA (Sharded, Isolated, Sliced, and Aggregated) framework for class-level deletion in CNN models, conducted on the CIFAR-10 dataset with the goal of reducing retraining time while maintaining model accuracy after removing entire classes. We began with a baseline CNN model trained on the complete dataset, where unlearning a class required full retraining on the reduced dataset, which proved computationally expensive and motivated the exploration of more efficient alternatives. The original SISA framework partitioned the dataset into multiple shards, with each shard training an independent constituent model, but the balanced class distribution across slices meant that class-level unlearning still required modifying all slices within the affected shard, limiting efficiency gains. To address this, we introduced Sequential Class-Level Slicing (SCLS), where distinct classes were assigned to different slices within each shard, enabling targeted slice removal and reducing the number of slices requiring retraining. However, this approach suffered from catastrophic forgetting, as the model lost knowledge of earlier classes during sequential training. To mitigate this issue, we incorporated a Reinforced Replay Training Mechanism (RRTM) that introduced 30\% of earlier samples during each slice's training phase, substantially improving accuracy by allowing the model to retain prior knowledge while learning new classes. Finally, we integrated a lightweight Gating Network that routes each input to the most appropriate constituent model based on learned shard representations, improving prediction accuracy by approximately 10\% compared to the aggregation-based approach while also reducing inference cost by activating only one constituent model per input rather than all models. The experimental results demonstrated that our final framework achieved competitive accuracy of 73.1\% with replay, further improved with the gating network, while significantly reducing retraining time compared to the baseline, with confusion matrix analysis confirming exact unlearning by showing zero predictions for the removed class after the unlearning process.

\section{Future Work}

Our research was limited by exploring the different modifications of SISA Framework. The objective of this research was to improve the architecture of the SISA framework for class level unlearning. Although our proposed modified SISA Framework is designed for robust datasets, we have only tasted it on the CIFAR 10 dataset. The next iteration of our experiments will include a larger dataset with high resolution images. CIFAR 10 is a balanced dataset. We will test the modified architecture with imbalanced datasets to evaluate the robustness of our proposed architecture. The next goal of architecture modification is to upgrade the current setup to handle datasets with overlapping class labels (i.e. COCO Dataset) and achieve efficient unlearning.

\bibliographystyle{IEEEtran}
\bibliography{references}

@article{Li2020A,title={A Survey of Convolutional Neural Networks: Analysis, Applications, and Prospects},author={Zewen Li and Fan Liu and Wenjie Yang and Shouheng Peng and Jun Zhou},journal={IEEE Transactions on Neural Networks and Learning Systems},year={2020},volume={33},pages={6999-7019},doi={10.1109/TNNLS.2021.3084827}}

@article{Santos2022Avoiding,title={Avoiding Overfitting: A Survey on Regularization Methods for Convolutional Neural Networks},author={C. F. G. Santos and J. Papa},journal={ACM Computing Surveys (CSUR)},year={2022},volume={54},pages={1 - 25},doi={10.1145/3510413}}

@article{badhon1_Bowling2006Machine,title={Machine learning and games},author={Michael Bowling and Johannes Fürnkranz and T. Graepel and R. Musick},journal={Machine Learning},year={2006},volume={63},pages={211-215},doi={10.1007/s10994-006-8919-x}}

@article{badhon3_Tax2017The,title={The Partial Information Decomposition of Generative Neural Network Models},author={T. M. S. Tax and P. Mediano and M. Shanahan},journal={Entropy},year={2017},volume={19},pages={474},doi={10.3390/e19090474}}

@INPROCEEDINGS{siam3_omi2_hemel1_ish2_9519428,

  author={Bourtoule, Lucas and Chandrasekaran, Varun and Choquette-Choo, Christopher A. and Jia, Hengrui and Travers, Adelin and Zhang, Baiwu and Lie, David and Papernot, Nicolas},

  booktitle={2021 IEEE Symposium on Security and Privacy (SP)}, 

  title={Machine Unlearning}, 

  year={2021},

  volume={},

  number={},

  pages={141-159},

  doi={10.1109/SP40001.2021.00019}}

@misc{siam4_liu2022continuallearningprivateunlearning,
      title={Continual Learning and Private Unlearning}, 
      author={Bo Liu and Qiang Liu and Peter Stone},
      year={2022},
      eprint={2203.12817},
      archivePrefix={arXiv},
      primaryClass={cs.AI},
      url={https://arxiv.org/abs/2203.12817}, 
}

@misc{siam8_fan2024salunempoweringmachineunlearning,
      title={SalUn: Empowering Machine Unlearning via Gradient-based Weight Saliency in Both Image Classification and Generation}, 
      author={Chongyu Fan and Jiancheng Liu and Yihua Zhang and Eric Wong and Dennis Wei and Sijia Liu},
      year={2024},
      eprint={2310.12508},
      archivePrefix={arXiv},
      primaryClass={cs.LG},
      url={https://arxiv.org/abs/2310.12508}, 
}

@INPROCEEDINGS{omi1_ish1_7163042,
  author={Cao, Yinzhi and Yang, Junfeng},
  booktitle={2015 IEEE Symposium on Security and Privacy}, 
  title={Towards Making Systems Forget with Machine Unlearning}, 
  year={2015},
  volume={},
  number={},
  pages={463-480},
  doi={10.1109/SP.2015.35}}

@INPROCEEDINGS{omi5_hemel3_9157084,
  author={Golatkar, Aditya and Achille, Alessandro and Soatto, Stefano},
  booktitle={2020 IEEE/CVF Conference on Computer Vision and Pattern Recognition (CVPR)}, 
  title={Eternal Sunshine of the Spotless Net: Selective Forgetting in Deep Networks}, 
  year={2020},
  volume={},
  number={},
  pages={9301-9309},
  doi={10.1109/CVPR42600.2020.00932}}

@misc{omi6_hemel2_neel2020descenttodeletegradientbasedmethodsmachine,
      title={Descent-to-Delete: Gradient-Based Methods for Machine Unlearning}, 
      author={Seth Neel and Aaron Roth and Saeed Sharifi-Malvajerdi},
      year={2020},
      eprint={2007.02923},
      archivePrefix={arXiv},
      primaryClass={stat.ML},
      url={https://arxiv.org/abs/2007.02923}, 
}

@misc{omi3_ish5_ginart2019makingaiforgetyou,
      title={Making AI Forget You: Data Deletion in Machine Learning}, 
      author={Antonio Ginart and Melody Y. Guan and Gregory Valiant and James Zou},
      year={2019},
      eprint={1907.05012},
      archivePrefix={arXiv},
      primaryClass={cs.LG},
      url={https://arxiv.org/abs/1907.05012}, 
}

@misc{ish6_henighan2020scalinglawsautoregressivegenerative,
      title={Scaling Laws for Autoregressive Generative Modeling}, 
      author={Tom Henighan and Jared Kaplan and Mor Katz and Mark Chen and Christopher Hesse and Jacob Jackson and Heewoo Jun and Tom B. Brown and Prafulla Dhariwal and Scott Gray and Chris Hallacy and Benjamin Mann and Alec Radford and Aditya Ramesh and Nick Ryder and Daniel M. Ziegler and John Schulman and Dario Amodei and Sam McCandlish},
      year={2020},
      eprint={2010.14701},
      archivePrefix={arXiv},
      primaryClass={cs.LG},
      url={https://arxiv.org/abs/2010.14701}, 
}

@misc{ish7_shokri2017membershipinferenceattacksmachine,
      title={Membership Inference Attacks against Machine Learning Models}, 
      author={Reza Shokri and Marco Stronati and Congzheng Song and Vitaly Shmatikov},
      year={2017},
      eprint={1610.05820},
      archivePrefix={arXiv},
      primaryClass={cs.CR},
      url={https://arxiv.org/abs/1610.05820}, 
}

@article{ish10_Tarun_2024,
   title={Fast Yet Effective Machine Unlearning},
   volume={35},
   ISSN={2162-2388},
   url={http://dx.doi.org/10.1109/TNNLS.2023.3266233},
   DOI={10.1109/tnnls.2023.3266233},
   number={9},
   journal={IEEE Transactions on Neural Networks and Learning Systems},
   publisher={Institute of Electrical and Electronics Engineers (IEEE)},
   author={Tarun, Ayush K. and Chundawat, Vikram S. and Mandal, Murari and Kankanhalli, Mohan},
   year={2024},
   month=sep, pages={13046–13055} }

@ARTICLE{hemel5_10488864,
  author={Xu, Jie and Wu, Zihan and Wang, Cong and Jia, Xiaohua},
  journal={IEEE Transactions on Emerging Topics in Computational Intelligence}, 
  title={Machine Unlearning: Solutions and Challenges}, 
  year={2024},
  volume={8},
  number={3},
  pages={2150-2168},
  doi={10.1109/TETCI.2024.3379240}}

@article{hemel6_Chundawat_2023,
   title={Zero-Shot Machine Unlearning},
   volume={18},
   ISSN={1556-6021},
   url={http://dx.doi.org/10.1109/TIFS.2023.3265506},
   DOI={10.1109/tifs.2023.3265506},
   journal={IEEE Transactions on Information Forensics and Security},
   publisher={Institute of Electrical and Electronics Engineers (IEEE)},
   author={Chundawat, Vikram S. and Tarun, Ayush K. and Mandal, Murari and Kankanhalli, Mohan},
   year={2023},
   pages={2345–2354} }

@misc{hemel7_yoon2023fewshotunlearningmodelinversion,
      title={Few-Shot Unlearning by Model Inversion}, 
      author={Youngsik Yoon and Jinhwan Nam and Hyojeong Yun and Jaeho Lee and Dongwoo Kim and Jungseul Ok},
      year={2023},
      eprint={2205.15567},
      archivePrefix={arXiv},
      primaryClass={cs.LG},
      url={https://arxiv.org/abs/2205.15567}, 
}

@inproceedings{siam8_kga_wang-etal-2023-kga,
    title = "{KGA}: A General Machine Unlearning Framework Based on Knowledge Gap Alignment",
    author = "Wang, Lingzhi  and
      Chen, Tong  and
      Yuan, Wei  and
      Zeng, Xingshan  and
      Wong, Kam-Fai  and
      Yin, Hongzhi",
    editor = "Rogers, Anna  and
      Boyd-Graber, Jordan  and
      Okazaki, Naoaki",
    booktitle = "Proceedings of the 61st Annual Meeting of the Association for Computational Linguistics (Volume 1: Long Papers)",
    month = jul,
    year = "2023",
    address = "Toronto, Canada",
    publisher = "Association for Computational Linguistics",
    url = "https://aclanthology.org/2023.acl-long.740/",
    doi = "10.18653/v1/2023.acl-long.740",
    pages = "13264--13276"
}

@article{lecun2015deep,
  title={Deep learning},
  author={LeCun, Yann and Bengio, Yoshua and Hinton, Geoffrey},
  journal={Nature},
  volume={521},
  number={7553},
  pages={436--444},
  year={2015},
  publisher={Nature Publishing Group}
}

\end{document}